\pdfoutput=1

\documentclass[letterpaper, 10 pt, conference]{ieeeconf}  

\IEEEoverridecommandlockouts                              

\usepackage{graphics} 
\usepackage{epsfig} 
\usepackage{mathptmx} 
\usepackage{times} 
\usepackage{amsmath} 
\usepackage{amssymb}  
\usepackage{url}
\usepackage{hyperref}

\title{\LARGE \bf
Detecting and Mapping Trees in Unstructured Environments with a Stereo Camera and Pseudo-Lidar
}

\author{Brian H. Wang, Carlos Diaz-Ruiz, Jacopo Banfi, and Mark Campbell$^{1}$
\thanks{This work was supported by the Office of Naval Research under the BRC grant N00014-17-1-2175 and the MURI grant N00014-17-1-2699.}
\thanks{$^{1}$The authors are with the Sibley School of Mechanical \& Aerospace Engineering, Cornell University, Ithaca, NY 14853, USA.
        {\tt\small \{bhw45\}, \{cad297\}, \{jb2639\}, \{mc288\} @cornell.edu}}%
}

\begin{document}

\maketitle
\thispagestyle{empty}
\pagestyle{empty}


\newcommand{\etal}{\emph{et al.}}
\newcommand{\githuburl}{\url{https://github.com/brian-h-wang/pseudolidar-tree-detection}}

\begin{abstract}
We present a method for detecting and mapping trees in noisy stereo camera point clouds, using a learned 3-D object detector. Inspired by recent advancements in 3-D object detection using a pseudo-lidar representation for stereo data, we train a PointRCNN detector to recognize trees in forest-like environments. We generate detector training data with a novel automatic labeling process that clusters a fused global point cloud. This process annotates large stereo point cloud training data sets with minimal user supervision, and unlike previous pseudo-lidar detection pipelines, requires no 3-D ground truth from other sensors such as lidar. Our mapping system additionally uses a Kalman filter to associate detections and consistently estimate the positions and sizes of trees. We collect a data set for tree detection consisting of 8680 stereo point clouds, and validate our method on an outdoors test sequence. Our results demonstrate robust tree recognition in noisy stereo data at ranges of up to 7 meters, on 720p resolution images from a Stereolabs ZED 2 camera. Code and data are available at \githuburl{}.
\end{abstract}

\section{INTRODUCTION}

Agile autonomous navigation in complex and unstructured environments requires a robot to plan ahead and process noisy sensor measurements, in order to determine the existence and locations of any obstacles in its planned path. Consider a quadrotor UAV flying through a forest---tree trunks and branches may appear in the quadrotor's flight path with little warning, and agile navigation therefore depends on the quadrotor's ability to use noisy sensor data to reason about the uncertain presence or absence of obstacles ahead.

Current state of the art approaches to agile autonomous navigation typically use either lidar \cite{mohta2018fast, mohta2018experiments} or stereo cameras  \cite{ryll2019efficient}. Lidar sensors, while accurate, are heavy, power-hungry, and expensive. Stereo cameras are more lightweight and less costly, but the quality of stereo depth information degrades significantly with range \cite{wang2019pseudo}. 
For this reason, mapping approaches designed for obstacle avoidance based on stereo cameras generally discard sensor measurements past a limited range, and/or limit the maximum robot speed in complex or crowded environments where sensor noise is more prevalent. Additionally, the computational cost of updating 3-D occupancy grid maps grows significantly as the maximum mapping range and/or grid resolution is increased, creating further difficulty for planning around obstacles in complex environments. 

\begin{figure}
    \centering
    \includegraphics[width=\columnwidth]{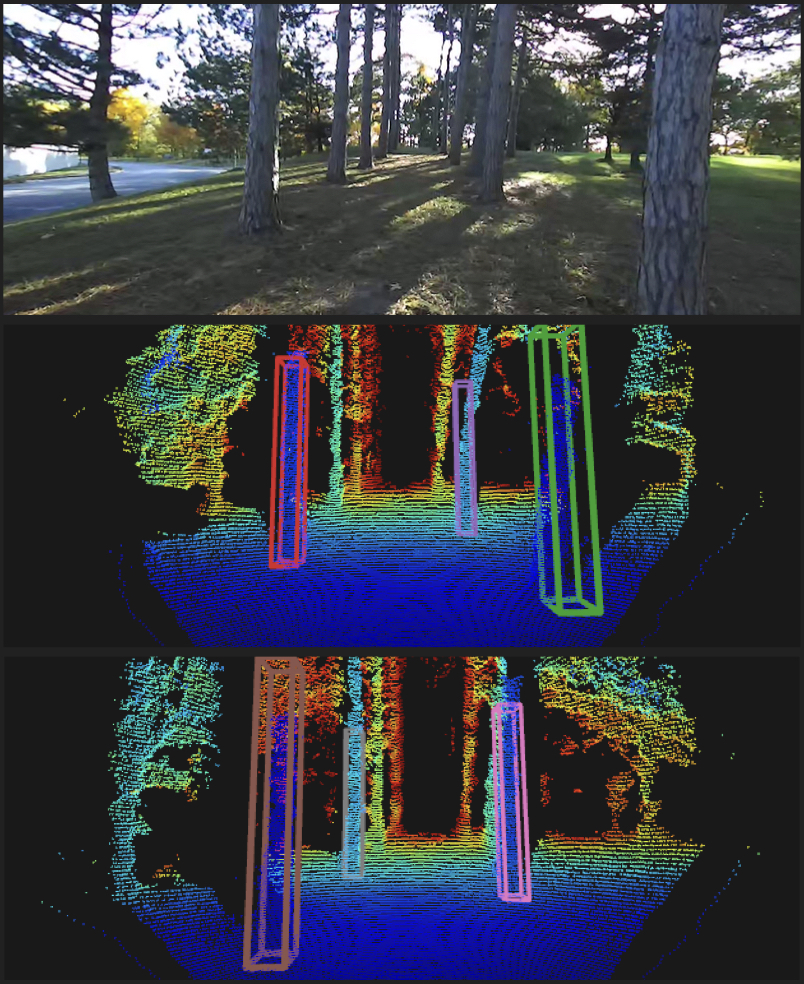}
    \caption{Results of our tree detection and mapping system, demonstrated on sample frames in our test sequence. The colored bounding boxes indicated different tracked objects within our tracker.}
    \label{fig:qual}
\end{figure}

Motivated by the problem of autonomous navigation in unstructured outdoors environments such as forests, and inspired by recent advancements in 3-D object detection using a \emph{pseudo-lidar} representation for stereo point clouds \cite{wang2019pseudo,you2019pseudo}, we present, to the best of our knowledge, the first method for detecting and mapping distant trees using a stereo camera alongside a state of the art 3-D object detector.  Figure \ref{fig:qual} demonstrates example results of our  method, which includes the following components:

\begin{enumerate}
    \item A labeling pipeline which generates a training data set for the tree detector from recorded stereo camera sensor data. 
    This pipeline allows rapid training data creation with minimal user effort by clustering a fused global point cloud map. 
    \item A PointRCNN 3-D object detector \cite{shi2019pointrcnn}, which detects trees in sparsified stereo point clouds, using the novel pseudo-lidar approach proposed in Wang \etal{} \cite{wang2019pseudo}. \label{step:detector} 
    \item A Kalman filter, which performs data association on detections in order to robustly map trees. The filter outputs accurate estimates of the positions and sizes of trees in front of the camera, and is scalable due to estimating a limited set of state variables per tree. \label{step:kalmanfilter} 
\end{enumerate}

\begin{figure}
    \centering
    \includegraphics[width=\columnwidth]{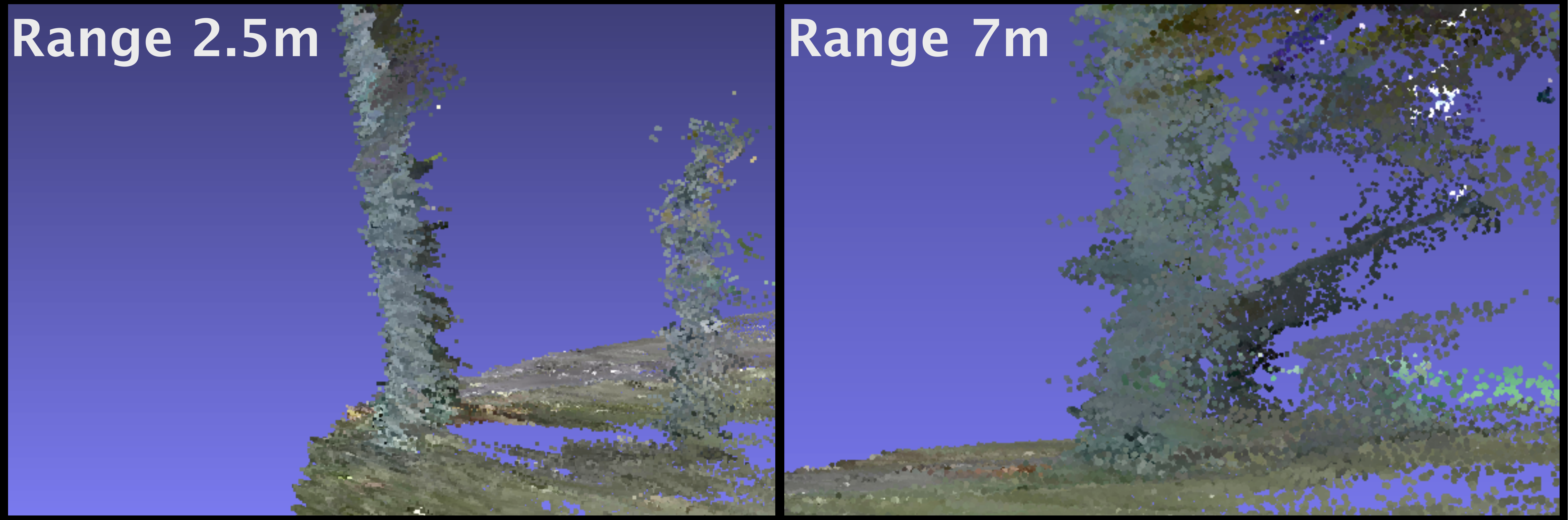}
    \caption{Example of the distortion that occurs in stereo point clouds at longer sensing ranges, using data from a ZED 2 stereo camera. \emph{Left:} Side view of a tree at a range of approximately 2.5 meters from the sensor, as seen in the 3-D point cloud generated from a stereo image pair. \emph{Right:} The same tree, at a range of 7 meters, appears significantly more distorted.}
    \label{fig:stereo_noise}
\end{figure}

By learning patterns from large numbers of labeled noisy stereo point clouds, our detection and mapping system is able to recognize trees that appear distorted in stereo data, an example of which is shown in Figure \ref{fig:stereo_noise}. We note that distorted objects such as this are infeasible to manually annotate with bounding box labels for training a 3-D detector, as there is no unbiased way for someone to judge the true position and size of the tree. For this reason, previous pseudo-lidar detection methods use training labels that come from annotated lidar data \cite{wang2019pseudo, you2019pseudo}. Our automatic labeling method avoids this problem, enabling the first stereo image training and detection pipeline with no dependence on lidar sensors. Our method is also modular, and can complement existing 3-D occupancy grid mapping approaches.

We validate our perception system on a test sequence of stereo point clouds captured in a forest-like environment, demonstrating accurate tree detection and mapping at ranges up to 7 meters away from the camera, where stereo depth noise becomes significant. To further promote creation of data sets and obstacle detectors for forests and other challenging environments for robotic navigation, our code for point cloud labeling, pseudo-lidar object detection, and tree mapping, as well as our collected data set, are publicly available at \githuburl{}.

\section{RELATED WORK}

\subsection{Perception for Navigation in Unstructured Environments}

Recent works have demonstrated robust and effective perception pipelines for enabling autonomous outdoors navigation. Ryll \etal{} \cite{ryll2019efficient} use an Intel RealSense D435i stereo camera to construct a 3-D occupancy grid that a UAV can use for navigation. However, the occupancy grid requires consistent sensor measurements to form a reliable map for planning; as a result, while the quadrotor could achieve peak speeds of 8 m/s in open areas, flight speed was greatly reduced around obstacles and narrow passageways. Mohta \etal{} \cite{mohta2018fast} use a similar perception component, the main difference being the usage of a nodding 2D lidar scanner as the main sensor for mapping. Due to computational considerations, lidar measurements are fused into a local occupancy grid map covering a distance only up to 7.5m away from the quadrotor. These navigation pipelines have demonstrated impressive results in autonomous quadrotor flight, however the difficulty of using longer-range sensor measurements remains a roadblock towards faster flight in more complex environments, with \cite{mohta2018fast} specifically identifying the size of the local map as a limiting factor on flight speed.

Other works have explored the problems of detecting trees and navigating in forests specifically. Giusti \etal{} \cite{giusti2015machine} use a convolutional neural network for detecting forest trails, allowing a quadrotor to navigate through the forest by visually following the trail. Mohta \etal{} \cite{mohta2018experiments} demonstrate quadrotor navigation in a sparse forest environment, and Tian \etal{} \cite{tian2018search} present a multi-UAV exploration framework using trees as landmarks for SLAM, clustering trees in order to reduce the computational burden of mapping a forest using an occupancy grid. Both of these pipelines use lidar sensing for mapping and tree detection.  Finally, Durand-Petiteville \etal{} \cite{durand2018tree} present a method for detecting tree trunks in orchards using stereo cameras, by reasoning about the shadows that tree trunks cast within the stereo point cloud. This method is computationally efficient and demonstrates that trees can be effectively recognized in stereo point clouds, however since it is designed specifically for usage in orchards, it depends on a number of heuristic assumptions about the shapes and sizes of the trees.

\subsection{3-D Pseudo-lidar Object Detection}

Motivated in large part by autonomous driving, researchers have made significant progress in perception and scene understanding with 3-D sensors \cite{yang2018pixor, milioto2019rangenet++, shi2019pointrcnn, kaul2020rss, shi2020pv}. Recent works have greatly narrowed the gap between 3-D object detectors that use lidar, and those that use stereo images as inputs. Wang \etal{} \cite{wang2019pseudo} showed that this gap was caused mainly by the representation used for stereo data, rather than the quality of depth estimation, as was previously assumed. By representing stereo depth images as a 3-D point cloud, then using the resulting point cloud as an input to a 3-D lidar object detector, they achieved massively improved detection accuracy over previous stereo image object detectors. Wang \etal{} refer to this representation as \emph{pseudo-lidar}, since the stereo point cloud mimics the measurements produced by a lidar sensor. You \etal{} \cite{you2019pseudo} introduced improvements that further increased the accuracy of stereo-based 3-D detection. Due to the reduced cost and weight of stereo cameras as compared to lidar, these advancements are especially impactful for UAVs and other resource-constrained robotic platforms.


\def\limg{I_{L,k}}
\def\rimg{I_{R,k}}


\section{APPROACH}\label{sec:approach}
Our approach to tree detection and tracking consists of three main components, described in detail in the following sections: a pipeline for generating training data (Section \ref{sec:approach-a}), a 3-D tree detector (Section \ref{sec:approach-b}), and a tree state estimator (Section \ref{sec:approach-c}).

\begin{figure*}[t]
    \centering
    \includegraphics[width=\textwidth]{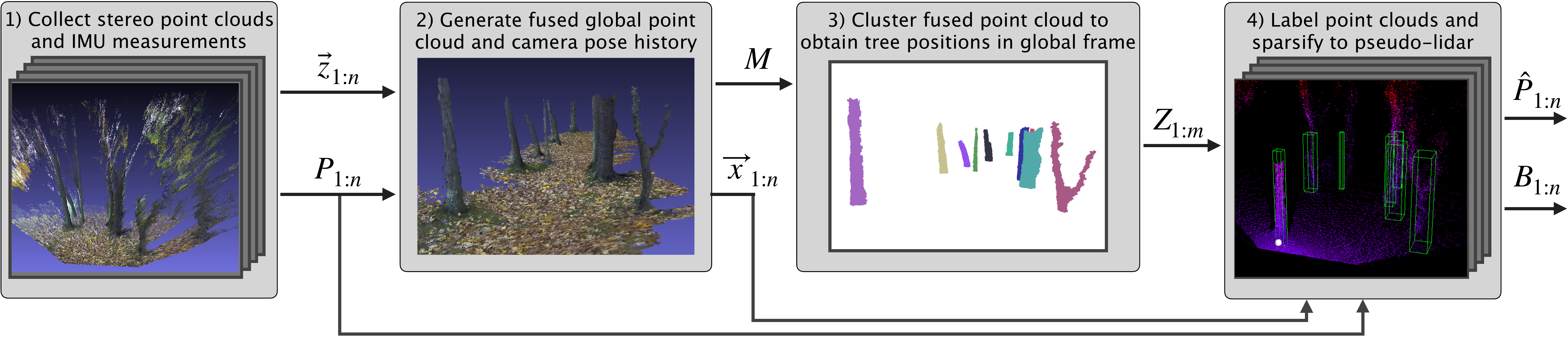}
    \caption{Our pipeline for generating labeled stereo point cloud training data for the 3-D tree detector. See section \ref{sec:approach-a} for definitions of terms.}
    \label{fig:pipeline}
\end{figure*}

\subsection{Training Data Generation Pipeline}

\label{sec:approach-a}

In the following section, we describe our pipeline for generating stereo image training data with minimal user supervision. Figure \ref{fig:pipeline} illustrates the steps of this process. We implement our pipeline in Python, using the Open3D \cite{Zhou2018open3d} library for 3-D data processing.

\subsubsection{Data collection}\label{subsec:data_collection}

Our approach uses these inputs:
\begin{itemize}
    \item $P_k$, point clouds from a 3-D camera sensor.
    \item $f_x, f_y, c_x, c_y$, the camera intrinsic parameters.
    \item $\vec{z}_k$, measurements from an IMU and any accompanying sensors used for camera pose estimation. These are used to ensure accurate mapping and localization, which are required for generating reliable training labels.
\end{itemize}

 For the camera point clouds and other sensor data, the index $k$ is over the time step indices in the data collection, $k\in\{1,\dots,n\}$. We use a Stereolabs ZED 2 stereo camera for our data collection.
 As described in \cite{wang2019pseudo}, given a pair of stereo images, the point cloud $P_k$ can be computed by matching pixels in the left image to corresponding pixels in the right image, then computing a depth value to each pixel based on the camera parameters.

We note that active RGBD sensors such as the Intel RealSense or Microsoft Kinect can also be used to obtain the required 3-D point cloud measurements and are therefore compatible with our pipeline. However, these sensors tend to perform less accurately outdoors due to their usage of infrared sensors. Thus, the ZED 2 stereo camera better addresses the use case we explore in this paper.

\subsubsection{Global point cloud fusion and pose estimation}\label{subsec:pcd_fusion}

Given the sensor measurements and camera parameters as inputs, we compute the following:
\begin{itemize}
    \item $\vec{x}_k$, the pose of the camera at each time step.
    \item $M \in \mathbb{R}^{p_{G} \times 3}$, a \emph{global point cloud map} created by fusing all $P_k$ into a global reference frame. $p_G$ is the total number of points in the global point cloud.
\end{itemize}

Multiple approaches for visual-inertial 3-D SLAM have been proposed, which generate the pose and point cloud map given a series of 3-D point clouds and inertial measurements \cite{labbe2019rtab, oleynikova2017voxblox, grinvald2019volumetric, rosinol2020kimera}. Any of these approaches can, in principle, be used to generate the required inputs for our labeling pipeline. Our current implementation uses the spatial mapping API of the Stereolabs ZED SDK\footnote{https://www.stereolabs.com/docs/spatial-mapping/}, which uses stereo vision alongside the built-in IMU, magnetometer, and barometer on the ZED 2 sensor to perform point cloud fusion and pose optimization.

\subsubsection{Global point cloud clustering}

After constructing the global point cloud, we apply a clustering algorithm in order to determine the number, shape, and position of individual tree trunks in the point cloud, proceeding as follows:

\begin{enumerate}
    \item Detect the ground plane in the global point cloud using RANSAC, and remove any points within a distance threshold to the ground.
    \item Cluster the remaining points using the DBSCAN clustering algorithm \cite{ester1996dbscan}. The output of this step is a list of clusters $Z_i \in \mathbb{R}^{p_i \times 3}$, where $p_i$ is the number of points in cluster $i$.
    \item Remove any cluster $i$ where $p_i < p_{\min}$, a threshold on cluster size. This filters out small erroneous clusters from sources such as low-hanging foliage.
    \item Output the remaining tree clusters $Z_i, i\in\{1,\dots,m\}$.
\end{enumerate}

This clustering approach is sensitive to noise in the 3-D points; thus, it cannot be used to effectively detect trees in the individual noisy stereo point clouds. However, noise in the fused global point cloud is significantly reduced in comparison, and the global cloud can therefore be accurately clustered to determine the number, shape, and position of individual tree trunks, as demonstrated in Figure \ref{fig:pipeline}.

In addition, we have developed a labeling tool which can be optionally used to review the generated clusters, remove any incorrect clusters, and manually annotate any trees missed by the DBSCAN clustering by drawing bounding boxes around groups of unclustered points in the global point cloud (with the ground plane removed). This step ensures the generated clusters are free of errors, which would otherwise propagate into the detector training data.

\subsubsection{Automated label generation and point cloud sparsification}

Given a cluster of 3-D points $Z_i$ for each tree in the global frame, we can step through the individual stereo point clouds $P_k$ and, using the computed pose of the camera $\vec{x}_k$ in the global frame, transform the clusters $Z_i$ to the local camera frame to determine which trees are visible at time $k$.

Our label generation procedure is as follows. For each time step $k\in \{1,\dots,n\}$:

\def\tmat{T_{W\rightarrow C, k}}
\def\pl{\hat{P}_k}

\begin{enumerate}
    \item Initialize the set of bounding box annotations for frame $k$ to $B_k = \emptyset$.
    \item Using the computed pose (position and orientation) of the camera $\vec{x}_k$, define the homogeneous transformation matrix $\tmat$ which transforms points from the global world frame to the local camera reference frame. 
    \item For each cluster of 3-D points $Z_i, i\in\{1,\dots,m\}$, use $\tmat$ to transform the cluster to its representation in camera frame coordinates, which we denote as $Z_i^C$.
    \item Fit a 3-D rectangular bounding box $b_{i,k}$, aligned to the axes of frame $C$, around the points $Z_i^C$. We align boxes to $C$, because otherwise the correct orientation of the box is undefined, since trees have no clear front side.
    \item If no points in the local point cloud $P_k$ fall within $b_{i,k}$, discard $b_{i,k}$. This removes trees which are outside the sensing range of the camera, or are completely occluded and therefore impossible to detect. Otherwise, add $b_{i,k}$ to the set $B_k$ of labels for point cloud $P_k$.
\end{enumerate}
In addition, following You \etal{} \cite{you2019pseudo}, we sparsify each stereo point cloud $P_k$ into a sparse pseudo-lidar representation $\pl$ by downsampling $P_k$ to 128 scan lines, evenly distributed according to vertical angle to the sensor, from $\theta=-35^\circ$ to $\theta=+35^\circ$ (covering the vertical field of view of the ZED 2 camera). We choose 128 scan lines because You \etal{} used 64 scan lines to imitate a lidar with a vertical field of view of approximately $30^\circ$; since the ZED 2 vertical field of view is $70^\circ$, doubling the number of pseudo-lidar scan lines maintains a roughly similar density of scan lines. The sparsification step is required because state of the art methods for 3-D object detection assume lidar point cloud inputs, and exploit the sparsity property of lidar data \cite{shi2019pointrcnn}. Sparsification has the added advantage of greatly reducing the size of the stereo point clouds---sparsifying ZED 2 stereo point clouds generated using 720p resolution stereo images results in an over tenfold decrease in the number of points---significantly reducing the storage requirements for a large data set. 

The end result of this pipeline is a set of sparse stereo point clouds $\pl$, each labeled with a set of ground truth bounding boxes $B_k$ indicating tree positions and sizes, to be used for 3-D object detector training. By automatically clustering a global point cloud, then optionally having a user spend a few minutes reviewing and fixing the clustering results, we can generate annotation bounding boxes for the hundreds or thousands of individual stereo point clouds used to create the fused global point cloud. We note that even when the user needs to add additional manual annotations, annotating a group of points in the ground plane-removed global point cloud takes only a few seconds of user time, and this single annotation is then used to label many tree instances in the individual sparse point clouds $\pl$.

Our approach assumes a forest environment, with all obstacles in the point cloud map being trees. In order to consider more general environments, with multiple object classes present, a potential extension to our pipeline is an extra labeling step where a user manually reviews each computed cluster and determines an appropriate class from a list of predefined options. This process would require little extra user time on top of our current pipeline, and would enable training multi-class 3-D object detectors.

\subsection{3-D Object Detector}

\label{sec:approach-b}

We use the PointRCNN detector, developed by Shi \etal{} \cite{shi2019pointrcnn} and publicly released online\footnote{https://github.com/sshaoshuai/PointRCNN}, to detect trees in 3-D stereo point clouds. Previously, You \etal{} \cite{you2019pseudo} applied PointRCNN to stereo point clouds, using the sparse pseudo-lidar representation. Our training data for the detector consists of the sparsified local stereo point clouds $\hat{P}_k$, as well as the 3-D tree bounding boxes $B_{i,k}$ generated through our labeling pipeline.

\subsection{Kalman Filter Estimator}

\label{sec:approach-c}

We treat 3-D bounding box detections from PointRCNN as uncertain measurements in a Kalman filter mapping framework, in order to consistently estimate tree positions and sizes and account for missed detections. Our goal is to estimate the joint probability $p(\mathbf{a}_k,\mathbf{t}_k|\mathbf{d}_k)$ at time $k$, where $\mathbf{t}_k$ is the set of estimated tree states (position and size), $\mathbf{d}_k$ is the set of observed sensor measurements (frame-wise bounding box detections), and $\mathbf{a}_k$ the assignment of measurements to mapped trees. We decouple the joint distribution into the continuous estimation problem $p(\mathbf{t}_k|\mathbf{a}_k,\mathbf{d}_k)$, solved recursively via a Kalman Filter (KF), and the discrete data assignment problem, $p(\mathbf{a}_k|\mathbf{t}_k,\mathbf{d}_k)$, solved via global nearest-neighbors (GNN).

\subsubsection{Kalman Filter}
We define a state vector for each tree based on its global frame position and size, 
\begin{equation}
\mathbf{t}^{i}_{k}=
    \begin{bmatrix}
    x & y & w 
    \end{bmatrix}^T,
\end{equation}
where $x, y$ is the position of the center of the tree in top-down birds-eye view coordinates, $w$ is the tree diameter, and $\mathbf{t}^{i}_{k}$ corresponds to the \textit{i}-th tree at the \textit{k}-th time. For the rest of the paper, the superscript \textit{i} is omitted for clarity; but it is given that multiple trees are being tracked. The $x$-axis points along the direction the camera is facing, and the $y$ axis points to the left. Our state vector effectively represents each tree as a cylinder, ignoring its height. We deem this an appropriate choice for representing tree trunks in our use case, as we would like to enable a robot to navigate between tree trunks in a forest, rather than flying over the tree canopies.

We assume the states of the model for each tree to be constant relative to the global frame, and driven by a small amount of process noise 
\begin{equation}
\mathbf{\dot{t}}_{k}=
    \begin{bmatrix}
    e_x & e_y & e_w
    \end{bmatrix}^T
\end{equation}
where $e_x, e_y, e_w$ are process noise values with small covariances which can be tuned to balance the time response of the filter and the pose error. This noise allows the state estimates to adjust over time as additional data arrives.

 The Kalman Filter prediction step is 
 \begin{equation}
    \mathbf{\hat{t}}_{k+1}=\mathbf{A}\mathbf{\hat{t}}_{k}
\end{equation}
\begin{equation}
    \hat{\Sigma}_{k+1}=\mathbf{A}\hat{\Sigma}_{k}\mathbf{A}^T+\mathbf{Q}
\end{equation}
 where $\mathbf{\hat{t}}$ is the state estimate, $\mathbf{A}$ is the state transition matrix (in this case identity), $\hat{\Sigma}$ is the state error covariance, and $\mathbf{Q}$ is the process noise covariance matrix.

Each measurement consists of direct measurements of all three states, derived from the bounding box detector. More specifically, we extract the following measurement vector $\mathbf{d}^m_{k+1}$ for the \textit{m}-th measurement at the \textit{k}$+1$ time: 
\begin{equation}
\mathbf{d}^m_{k+1}=
    \begin{bmatrix}
    d_x & d_y & d_w
    \end{bmatrix}^T,
\end{equation}
where $d_x,d_y$ denote the center-bottom position of the bounding box relative to the camera coordinate system, and $d_w$ is its width. We treat the width of a bounding box as a noisy measurement of the corresponding tree's diameter. The measurement function from the detections $h_{\mathrm{d}}(\cdot)$ is
\begin{equation}
    \mathbf{d}_{k+1}=\mathbf{H}\mathbf{t}_{k+1}+\mathbf{r}_{k+1}.
\end{equation}
The measurement function is modeled with additive zero-mean Gaussian white noise $\mathbf{r}_{k+1}$. This yields the predicted measurement and corresponding innovation covariance to be:
\begin{equation}
    \mathbf{\hat{d}}_{k+1}=\mathbf{H}\mathbf{\hat{t}}_{k+1}
\end{equation}
\begin{equation}
    \mathbf{S}_{k+1}=\mathbf{H}\Sigma_{k+1}\mathbf{H}^T+R.
\end{equation}
The tree location and diameter estimates are then updated according to the standard Kalman Filter update equations.

\subsubsection{Data Association}
 We use the GNN algorithm to optimally assign measurements to tracks, using the Hungarian algorithm~\cite{Assignment} to minimize the Mahalanobis distance 
    \begin{equation}
        D^{j,i}_{k+1} = (\mathbf{d}_{k+1}^j - \hat{\mathbf{d}}_{k+1}^i)^T S^{-1}_{k+1} (\mathbf{d}_{k+1}^j - \hat{\mathbf{d}}_{k+1}^i) ,
    \end{equation}
where the \textit{j}-th measurement is compared to the \textit{i}-th estimated measurement, derived from the \textit{i}-th track, and $S$ is the corresponding innovation covariance matrix. 

Tracks are initialized when a fixed number $a_{\mathrm{minHits}}$ of measurements are associated to the same track; our current implementation has $a_{\mathrm{minHits}}=3$. Track deletion, on the other hand, requires that a current track does not receive any measurement updates over a set amount of time steps $a_{\mathrm{maxAge}}$, or that the object exits the field of view (FOV); we use $a_{\mathrm{maxAge}}=100$. For our 60FPS camera, this equates to $1.67$ seconds. This method of track deletion addresses short temporary occlusions. We use $\chi^2$ hypothesis testing for validation gating, where very unlikely ($<$5\%) measurement associations are discarded as clutter or false positives.

\section{EXPERIMENTS AND RESULTS}

\subsection{Data collection and processing}

In order to train our tree detection system, we collected data in forest-like environments near the Cornell University campus, using a handheld Stereolabs ZED 2 camera to gather over 40,000 stereo image pairs at 720p resolution, split up into seven individual sequences. Due to the camera's 60 FPS framerate, in order to avoid including multiple highly similar frames in our detector training data, we downsampled the data to only include every fifth frame in our training set, for a set of 8680 stereo point clouds in total. In terms of number of frames, this is similar to the popular 3-D object detection KITTI data set, which includes 7481 training frames \cite{geiger2012kitti}.

Each of the seven data capture sequences was fused into a separate global point cloud, which were each then clustered using our automatic annotation pipeline. For the ground plane removal, we ran RANSAC for 1000 iterations, and removed any points within 0.5 m of the fit plane. For the global point cloud clustering, we set the DBSCAN parameters to $\epsilon=0.1$ and $minSamples=10$. Finally, after DBSCAN clustering, we remove any clusters containing less than 2000 points (To give a sense of scale, the full global point clouds from our seven training data captures each contained between 1.1 and 2.2 million points, though the majority of these would be removed with the ground plane). When creating the sparsified pseudo-lidar point clouds, we also remove stereo points past 15 meters from the camera, to avoid creating training bounding boxes that include only very few 3-D points.

Our training data sequences contain 110 distinct trees in total. 98 of these were correctly automatically clustered by our global point cloud clustering process. The automatic clustering generated 5 incorrect clusters which a manual annotator removed. Finally, the manual annotator added 12 clusters which were missed by the DBSCAN clustering---most of these were very thin tree trunks which fell under the minimum threshold for DBSCAN cluster sizes. In total, we trained PointRCNN on 8680 pseudo-lidar point clouds annotated with 34,386 bounding box training labels.

\subsection{Evaluating tree detection and estimation accuracy}

\begin{figure}[t!]
    \centering
    \includegraphics[width=\columnwidth]{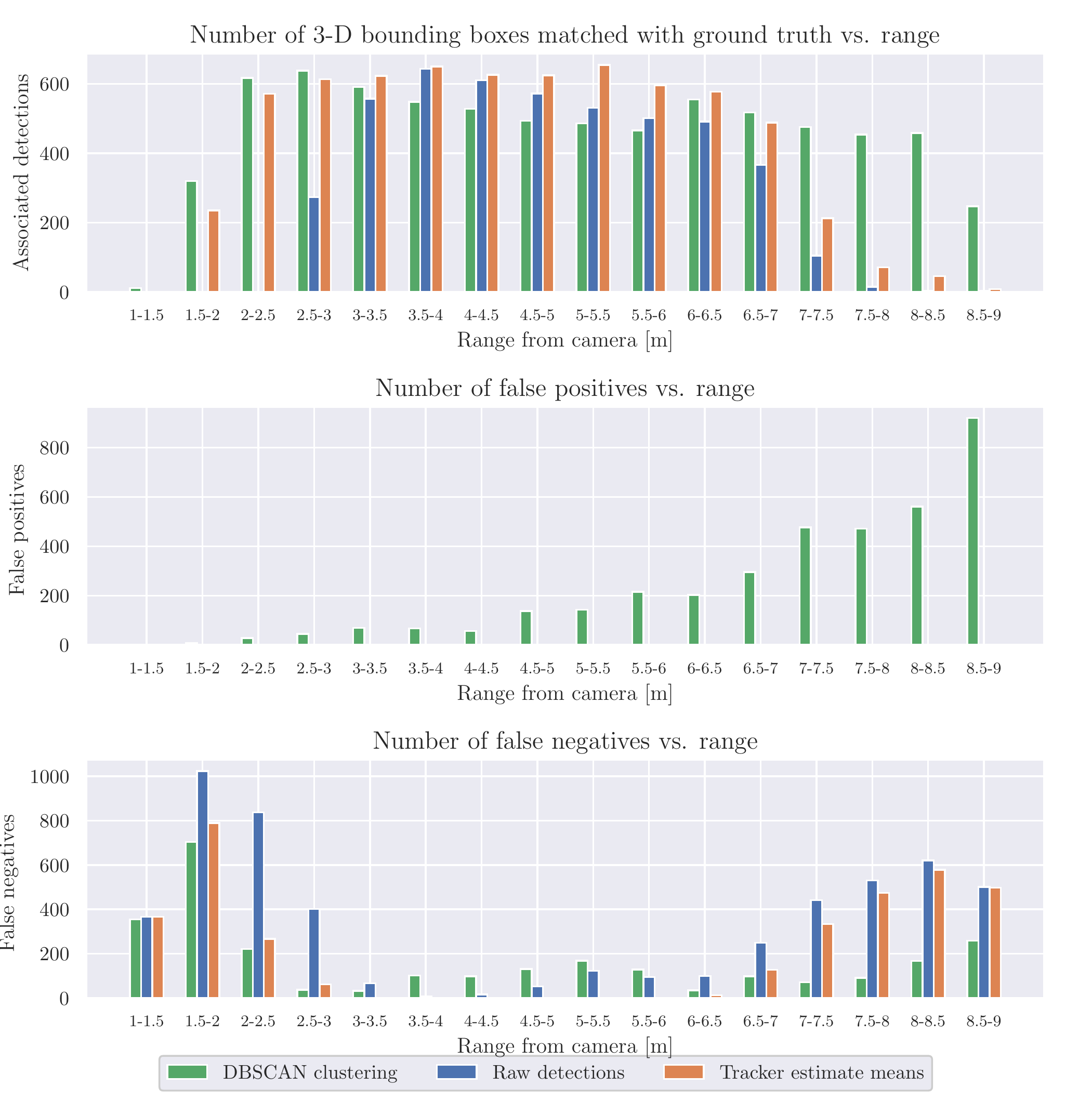}
    \caption{Counts of successfully matched 3-D bounding boxes, false positives, and false negatives, shown binned over different ranges from the sensor. We compare raw detector bounding box outputs and tracker estimates with baseline detections generated by running DBSCAN clustering on noisy stereo point clouds.}
    \label{fig:fp_fn_counts}
\end{figure}

\begin{figure*}[t!]
  \centering
 \includegraphics[width=2.3in]{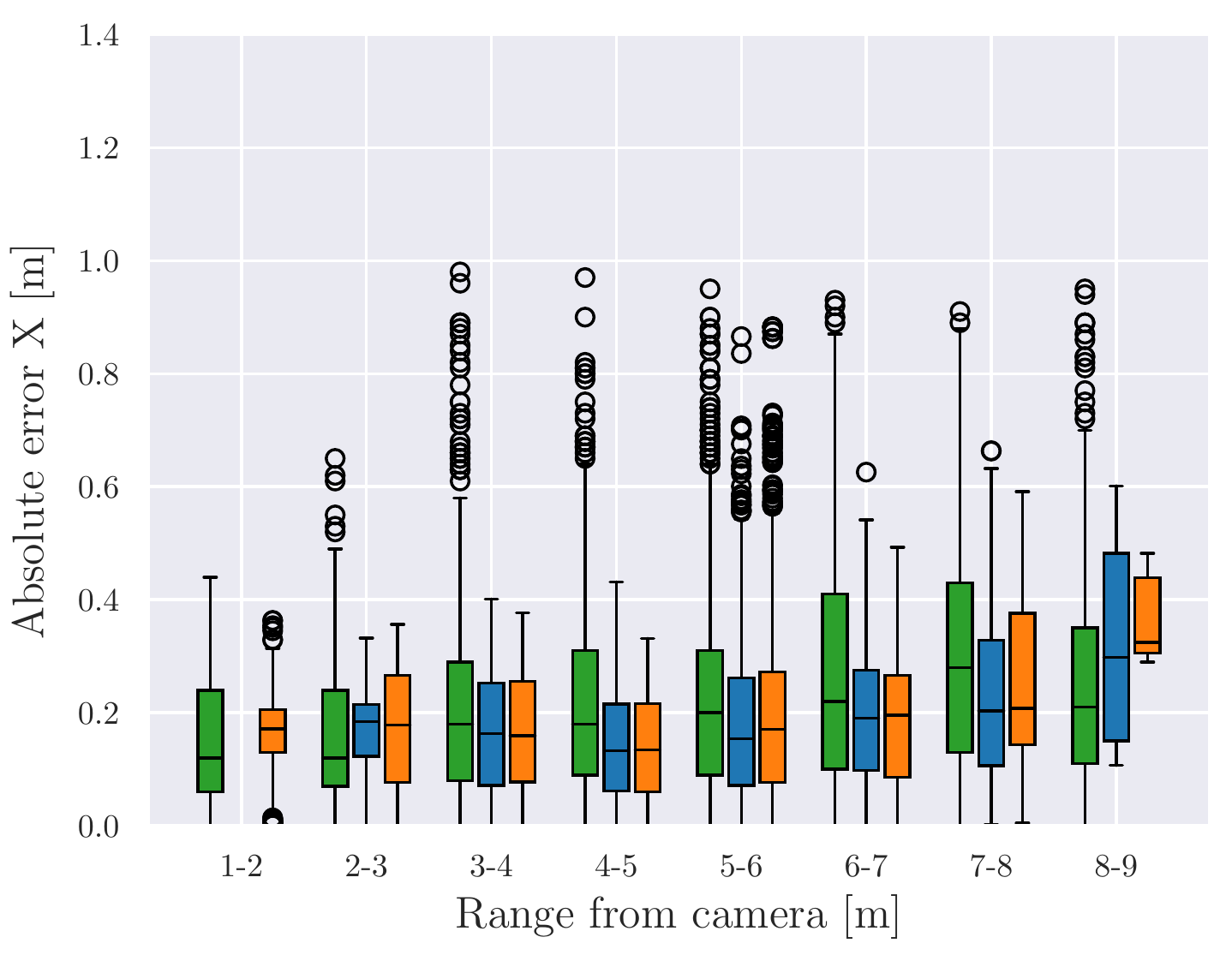}
 \includegraphics[width=2.3in]{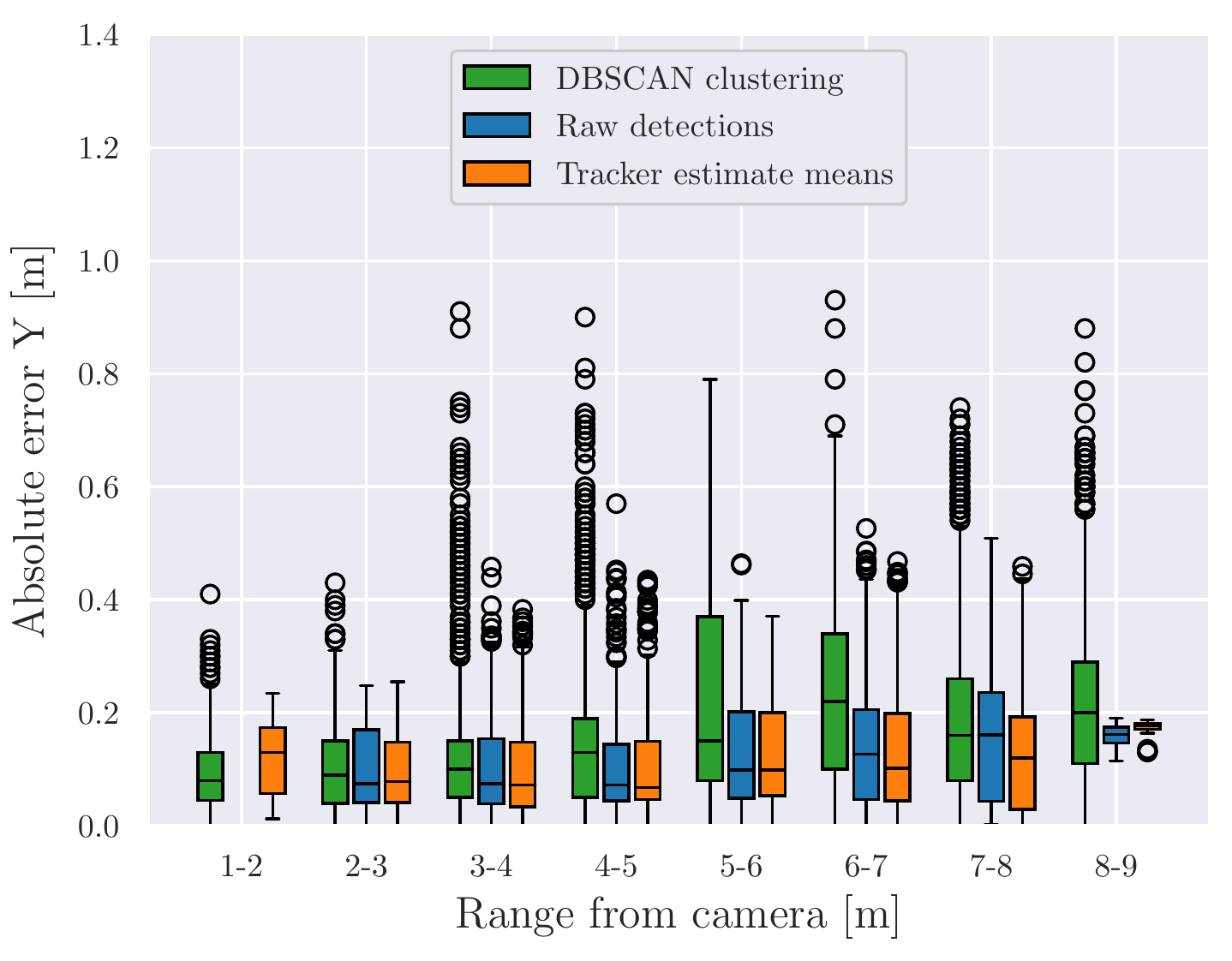}
 \includegraphics[width=2.3in]{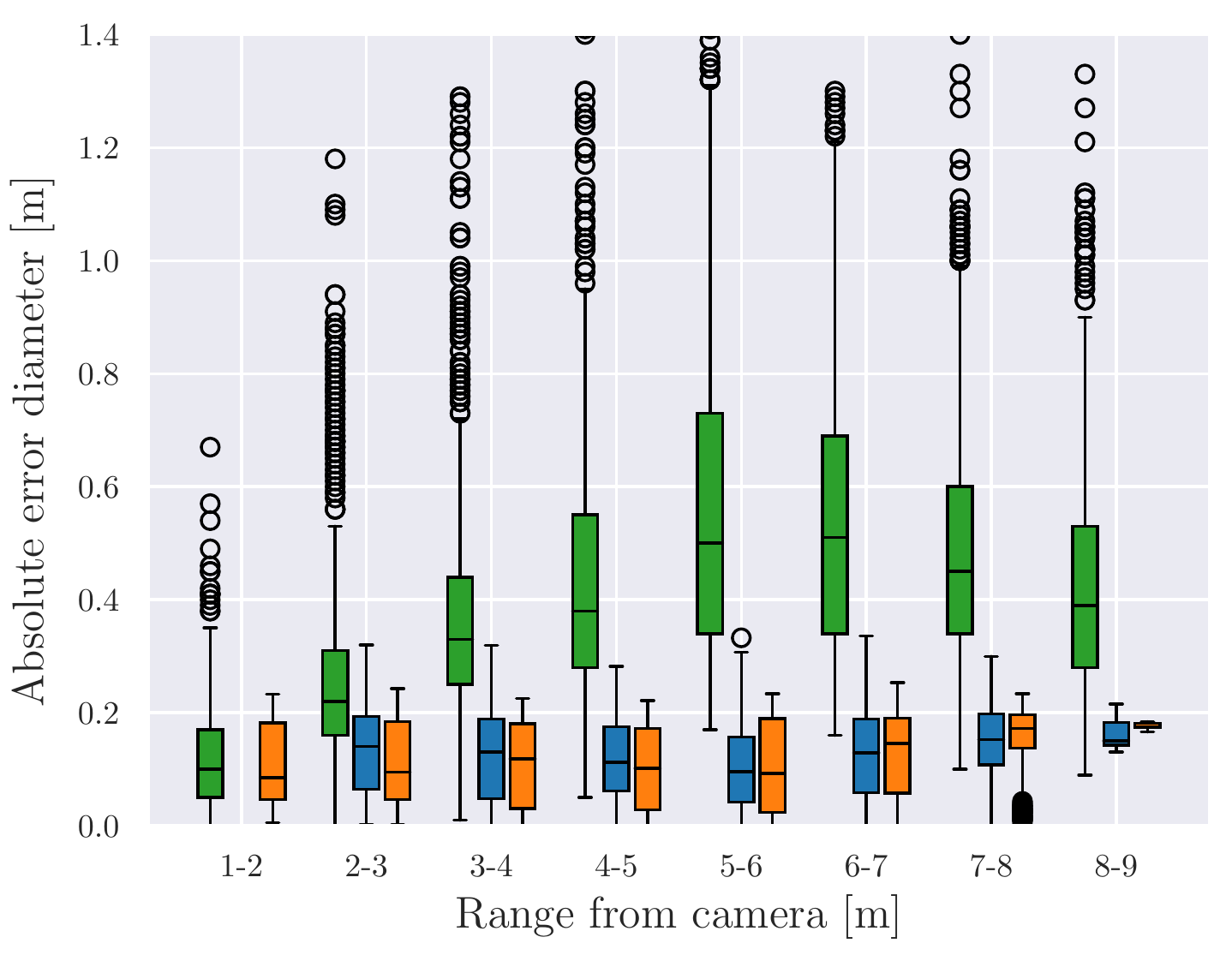}
\caption{Absolute errors on $x$ (forward axis) position, $y$ (lateral axis) position, and tree diameter over different ranges. 
An absent bar indicates no bounding boxes present over a given range bin.}
\label{fig:errors}
\end{figure*}

We evaluate our trained PointRCNN tree detector and tracker on a test sequence of 2000 stereo point clouds, captured in a separate forest-like environment. This site contains two rows of trees, and is physically separated from the area where we collected our training data, in order to evaluate our detector's performance in an environment outside the training domain. Figure \ref{fig:qual} visually shows our test site; for full results of our tree mapping system, please see the video accompanying this paper. We create ground truth tree annotations of this outdoors test sequence using our clustering and bounding box generation process, additionally using our manual annotation tool to ensure correct labels.

As a baseline for comparison, we run our ground plane removal and DBSCAN clustering method directly on the individual stereo point clouds $P_k$, fit a bounding box around each DBSCAN cluster, and treat these as detections. This baseline exactly mimics our clustering procedure used for automatic training data generation, except applied on noisy stereo point clouds rather than on the fused global point cloud. We leave all parameters used for this baseline clustering identical to the parameters used for generating training labels for the detector, except for running RANSAC ground plane removal for only 50 iterations (this saves computation time and did not significantly affect clustering, since the ground area in $P_k$ is much smaller than in the global point cloud and therefore does not require as precise a plane fit).

Figure \ref{fig:fp_fn_counts} shows the total number of bounding boxes successfully associated with ground truth boxes (top), false positive bounding boxes (middle), and false negatives (bottom) for the baseline DBSCAN detections, tree detector and the tracker. For evaluation, we match bounding boxes to nearby ground truth instances using linear assignment, with a maximum distance cutoff of 1 meter. This evaluation analyzes how often our detection and mapping system successfully recognizes the presence of trees ahead. We bin statistics over various ranges, in order to analyze how performance changes with distance from the sensor. As seen in the results, the DBSCAN baseline correctly recognizes the majority of ground truth trees at shorter ranges, when depth errors are minimal. Past around 5 meters, however, the number of false positive DBSCAN bounding boxes grows rapidly, as DBSCAN begins to mistakenly find clusters in noisy areas of the stereo point clouds. In contrast, our detector, trained on data generated largely by the same DBSCAN clustering method applied to a fused global point cloud, does not raise these false postives, and demonstrates a low false negative rate up to around 7 meters, significantly further than the accurate range of DBSCAN point cloud clustering. Past 5 meters, the Kalman filter effectively compensates for missed detections, reducing the false negative rate of the system. Our detector noticeably misses trees at very short ranges; at close distances, the stereo point cloud is very dense, which may make shorter-range detections difficult for PointRCNN, which was designed with lidar data in mind, to learn. This is however a minor drawback, since at such a close range depth measurements are quite reliable and could be easily integrated into a local grid-based map.

In order to examine the quality of tree bounding boxes, Figure \ref{fig:errors} reports the errors in $x$- and $y$-position and tree size for the DBSCAN baseline bounding boxes, PointRCNN detections, and Kalman filter estimates. Errors for both the PointRCNN detections and filtered tree states remain consistently low over the ranges shown, indicating that the detections closely match the ground truth positions and sizes of trees in the test sequence. In contrast, the DBSCAN baseline demonstrates low error up to around 3 meters, but begins to drastically grow in error after this point, with errors in tree sizes becoming particularly large.

\section{CONCLUSION}

We present a method for detecting and mapping trees in stereo camera point clouds. In our experiments, this system accurately recognizes trees at ranges up to 7 meters, even when trained on labels generated from clustering which is accurate in noisy stereo point clouds only up to approximately 3-4 meters. Our automatic labeling pipeline enables a detector to learn to recognize trees in noisy longer-range stereo points, which are often discarded in current mapping approaches, increasing the extent of usable sensor information available to the robot. Additionally, our pipeline bypasses the need for time-consuming manual labeling, and does not depend on supervision from a lidar sensor.  Future work will focus on further extending the range of learned 3-D object detectors, enabling robots to navigate while accounting for distant uncertain obstacles.

\addtolength{\textheight}{-10.8cm}   




\section*{ACKNOWLEDGMENT}

The authors would like to thank Yan Wang for helpful discussion on training the pseudo-lidar 3-D object detector.


\bibliographystyle{IEEEtran}
\bibliography{references}

\end{document}